\documentclass{article}

\usepackage{microtype}
\usepackage{graphicx}
\usepackage{subcaption}
\usepackage{booktabs} 
\usepackage{balance}
\usepackage{hyperref}

\usepackage[accepted]{icml2026}

\usepackage{amsmath}
\usepackage{amssymb}
\usepackage{mathtools}
\usepackage{amsthm}
\usepackage{multirow}

\usepackage[capitalize,noabbrev]{cleveref}

\theoremstyle{plain}
\newtheorem{theorem}{Theorem}[section]

\theoremstyle{definition}

\theoremstyle{remark}
\newtheorem{remark}[theorem]{Remark}

\usepackage[textsize=tiny]{todonotes}

\usepackage{makecell}
\usepackage{booktabs}
\usepackage{amsmath, amssymb}
\usepackage{xcolor}
\usepackage{amsthm}
\usepackage{adjustbox}
\usepackage{natbib}
\usepackage{tcolorbox}
\definecolor{deeppurple}{HTML}{673AB7}

\newcommand{\sd}[1]{{\tiny$\pm$#1}}
\usepackage{pifont}
\newcommand{\cmark}{\ding{51}}
\newcommand{\rone}[1]{{\color{red}\textbf{#1}}}
\newcommand{\rtwo}[1]{{\color{purple}\textbf{#1}}}
\newcommand{\rthree}[1]{{\color{blue}\textbf{#1}}}
\usepackage{tikz-cd}

\icmltitlerunning{Deep Neural Sheaf Diffusion}

\begin{document}

\twocolumn[
  \icmltitle{Deep Neural Sheaf Diffusion
}



  \icmlsetsymbol{equal}{*}

  \begin{icmlauthorlist}
    \icmlauthor{Rémi Bourgerie}{yyy}
    \icmlauthor{\v{S}ar\={u}nas Girdzijauskas}{yyy}
    \icmlauthor{Viktoria Fodor}{yyy}

  \end{icmlauthorlist}

  \icmlaffiliation{yyy}{School of Electrical Engineering and Computer Science, KTH Royal Institute of Technology, Stockholm, Sweden}

  \icmlcorrespondingauthor{Rémi Bourgerie}{remibo@kth.se}

  \icmlkeywords{Machine Learning, ICML}

  \vskip 0.3in
]

\printAffiliationsAndNotice{}

\begin{abstract}

Deep Graph Neural Networks (GNNs) are essential for
capturing complex dependencies in graph-structured data. However, scaling GNNs to depth remains challenging, as stacking layers leads to representation collapse and diminishing sensitivity due to repeated aggregation.
While Neural Sheaf Diffusion (NSD) provides strong theoretical guarantees against such collapse, 
these guarantees do not translate to practice: as depth increases, the disagreement signal of the sheaf Laplacian vanishes, limiting the contribution of deeper layers.
We identify mechanisms that hinder NSD effectiveness at depth and propose \emph{Deep Neural Sheaf Diffusion} (\textsc{DNSD}), which replaces the sheaf Laplacian with a sheaf adjacency operator to maintain informative signals across layers. This is complemented by normalization, odd nonlinearities, and gating.
To provide a principled explanation of the expected performance improvement, we contrast sheaf diffusion to graph attention mechanisms, highlighting that \textsc{DNSD} replaces scalar attention scores with matrix-valued edge functions and normalizes node representations rather than attention scores.
We demonstrate empirically that \textsc{DNSD} effectively utilizes deep aggregation in graph tasks, outperforming GNN and NSD baselines with up to 30pp accuracy on synthetic long-range datasets, and consistently outperforming them on real-world benchmarks.
These results position sheaf-based architectures as a promising building block for graph foundation models by supporting effective deep architectures.
Code available at: \url{https://github.com/remibourgerie/deep-neural-sheaf-diffusion}.

\end{abstract}

\section{Introduction}
\label{sec:intro}

Learning expressive representations on graphs requires architectures that propagate and transform information across multiple layers. However, training deep architectures is inherently difficult due to issues such as vanishing updates and unstable optimization. In domains
such as language and vision, these challenges have been mitigated by architectural design principles including residual connections, normalization, and gating~\citep{he2016deep,ba2016layer,qiu2025gated}. In contrast, how to achieve similar depth scaling in GNNs remains far less understood~\citep{bechler2026billion}.

Beyond these general challenges of deep models, GNNs exhibit a distinct difficulty arising from the message passing paradigm itself. Each layer aggregates information from neighbors through weighted averaging, and increasing depth results in repeated aggregation across the graph. In most message passing architectures, this aggregation takes the form of a convex combination of neighbor representations~\citep{kipf2017semi,velivckovic2017graph}. As this process is composed across layers, node representations become progressively less distinguishable, a phenomenon known as oversmoothing~\citep{chen2020measuring}. At the same time, repeated aggregation reduces the sensitivity of node representations to individual inputs, a phenomenon often referred to as oversquashing~\citep{di2023over}.
These effects reflect a deeper limitation: standard message passing is constrained to convex aggregation, which progressively reduces both representation diversity and sensitivity as depth increases. This suggests that the challenge is not only to stabilize deep architectures, but to move beyond convex aggregation as the interaction mechanism between nodes.

This limitation directly impacts the development of graph foundation models, which rely on depth to capture long-range structure and learn expressive, generalizable representations across tasks and datasets~\citep{wang2025graph}. Improving depth scaling is therefore a necessary architectural step toward graph foundation models. 

To address this bottleneck, a natural direction is to move beyond scalar aggregation and generalize how information is combined across edges.
Network sheaves~\citep{curry2014sheaves} provide a principled topological framework for modeling information flow on graphs by equipping edges with linear maps, allowing interactions to depend on richer local structure. 

In contrast to standard message passing, where messages are scalar-weighted by the graph structure or attention, \emph{Neural} Sheaf Diffusion (NSD)~\citep{bodnar2022neural} exploits learned, edge-specific linear maps to transform node representations along edges, through a non-linear diffusion process governed by the sheaf Laplacian.

These richer edge interactions lead to strong theoretical guarantees: in the linear setting, given an appropriate sheaf, the stationary distribution of sheaf diffusion can separate almost any label assignment, preventing representation collapse under repeated aggregation~\citep{bodnar2022neural}. These results suggest that sheaf-based architectures provide a promising foundation for building deep and expressive GNNs.

However, despite these guarantees, existing NSD architectures do not fully realize this potential in practice. We identify several mechanisms that limit their scalability to depth. Most importantly, the sheaf Laplacian measures disagreement between connected nodes. As diffusion reduces this disagreement, the update signal progressively vanishes, causing deeper layers to receive diminishing inputs. Beyond this signal degeneration, repeated composition introduces additional challenges. First, the scale of representations can vary across layers, leading to instability as depth increases. Second, standard nonlinearities such as ReLU introduce an asymmetric truncation, suppressing negative values under repeated composition. Finally, repeated aggregation propagates all contributions uniformly, allowing noise to accumulate across layers.

In this paper we address these challenges by proposing \textsc{DNSD}, a sheaf diffusion architecture that scales to depth, realizing the potential of neural sheaf diffusion. Our main contributions are as follows:

\begin{enumerate}

\item We propose DNSD, a modified sheaf diffusion architecture for deep GNNs. \textsc{DNSD}  replaces the sheaf Laplacian with a sheaf adjacency operator, shifting from measuring difference to aggregating dependency between neighboring representations. This is complemented by normalization, odd nonlinearities, and gating.

\item We contrast sheaf diffusion with graph attention mechanisms, highlighting that \textsc{DNSD} uses matrix-valued edge functions instead of scalar attention scores, and normalizes node representations rather than attention scores. 
These differences explain why \textsc{DNSD} maintains informative signals
under deep aggregation.

\item We evaluate \textsc{DNSD} on a synthetic benchmark designed to probe depth scaling, as well as on real-world datasets. We show that \textsc{DNSD} improves performance 
in deep architectures, and remains effective as depth increases, with the adjacency formulation as the dominant factor. These results suggest that adjacency-based sheaf diffusion is a promising building block for graph foundation model architectures that scale in depth.

\end{enumerate}

The rest of the paper is organized as follows. Section~\ref{sec:background} introduces the necessary background on sheaf diffusion. Section~\ref{sec:method} presents the proposed \textsc{DNSD} architecture. Section~\ref{sec:attention} discusses its connection to attention mechanisms. Section~\ref{sec:experiments} reports empirical results, and Section~\ref{sec:conclusion} concludes the paper.

\begin{figure}
    \centering
    \includegraphics[width=0.8\linewidth]{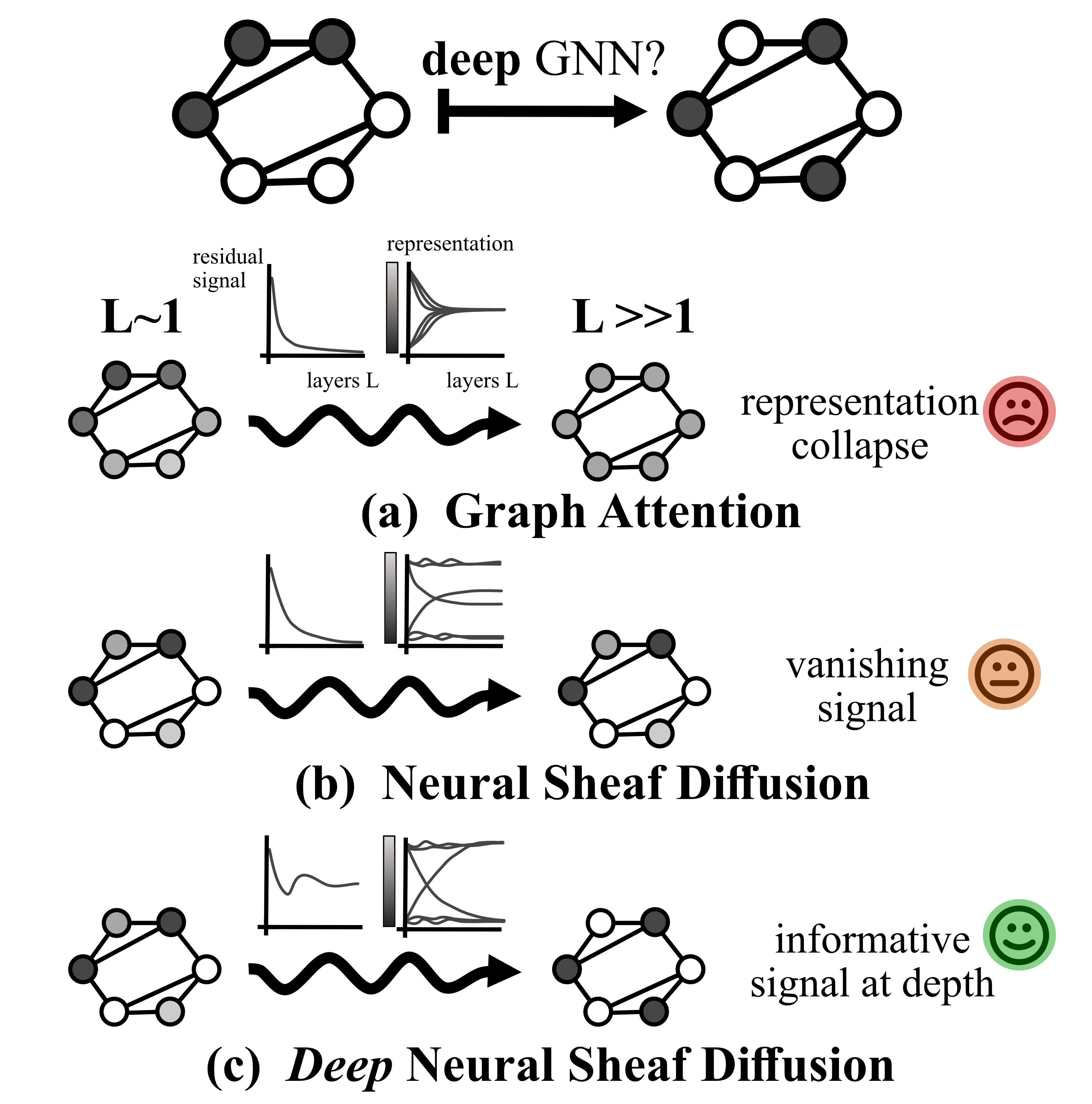}
    \caption{As depth increases ($L \gg 1$), standard message passing mechanisms degrade: graph attention (a) collapses representations, while neural sheaf diffusion (b) produces vanishing signals that limits the effective depth. Deep Neural Sheaf Diffusion (c) maintains informative signal at depth.}
    \label{fig:illustration}
\end{figure}

\section{Background and Related Work}
\label{sec:background}

\paragraph{Cellular sheaves on graphs.}
Sheaves provide a framework for describing how local data defined on a space can be consistently combined into global structure \citep{bredon1997sheaf}.
Cellular sheaves adapt this framework to discrete structures such as graphs~\citep{curry2014sheaves}, by associating vector spaces to nodes and linear maps to edges.
In this work, we restrict to real vector spaces, yielding a linear-algebraic formulation amenable to computation. Accordingly, a cellular sheaf $C(\mathcal{G}, \mathcal{F})$ on a graph $\mathcal{G} = (\mathcal{V}, \mathcal{E})$ with $n$ vertices is specified by the triple
$\langle \mathcal{F}(v), \mathcal{F}(e), \mathcal{F}_{v \trianglelefteq e} \rangle$, where for each $v \in \mathcal{V}$ and $e \in \mathcal{E}$,
$\mathcal{F}(v) = \mathbb{R}^{d}$ are node stalks,
$\mathcal{F}(e) = \mathbb{R}^{d}$ are edge stalks, and
$\mathcal{F}_{v \trianglelefteq e}: \mathcal{F}(v) \to \mathcal{F}(e)$ are linear restriction maps for each incident node--edge pair. In contrast to standard message passing in GNNs, where interactions are scalar-weighted by the graph, sheaf-based models allow node representations to be transformed along edges through learned linear maps.

\paragraph{Sheaf diffusion.}
This structure induces a sheaf Laplacian $L_\mathcal{F}$, which generalizes the graph Laplacian and quantifies disagreement between neighboring node representations after transformation by the edge maps~\citep{hansen2019toward}. Acting on stacked stalk signals $\mathbf{X} \in \mathbb{R}^{nd}$, it admits a block matrix representation with diagonal blocks $(L_\mathcal{F})_{uu} = \sum_{v \in \mathcal{N}(u)} \mathcal{F}_{u \trianglelefteq e}^\top \mathcal{F}_{u \trianglelefteq e}$ and off-diagonal blocks $(L_\mathcal{F})_{uv} = -\mathcal{F}_{u \trianglelefteq e}^\top \mathcal{F}_{v \trianglelefteq e}$ for $(u,v) \in \mathcal{E}$.
Let $\Delta_\mathcal{F} = D_\mathcal{F}^{-1/2} L_\mathcal{F} D_\mathcal{F}^{-1/2}$ denote the normalized sheaf Laplacian. A linear diffusion process then iteratively reduces disagreement as
\begin{equation}
\mathbf{X}^{(t+1)} = \left(I_{nd} - \Delta_\mathcal{F}\right) \mathbf{X}^{(t)}.
\label{eq:sheaf_diffusion}
\end{equation}
As $t \to \infty$, this process converges to $\ker(\Delta_\mathcal{F})$, corresponding to globally consistent sections. \citet{bodnar2022neural} show that, given appropriate restriction maps, any node classification task can be reduced to achieving sufficient separability in this space. This property suggests that, in the linear setting, sheaf diffusion may avoid representation collapse induced by repeated aggregation.

\paragraph{Neural Sheaf Diffusion (NSD).}
In practice, the diffusion steps are finite and the restriction maps are learned from data. At each layer, they are computed from the current node embeddings, making the sheaf layer-dependent. \citet{bodnar2022neural} parameterize this process as a neural architecture.
Node features are first mapped to stalk representations $\mathbf{X}^{(0)} \in \mathbb{R}^{nd \times f}$, where $d$ is the stalk dimension and $f$ the feature width per stalk. For each node $v$, we denote $\mathbf{X}_v^{(l)} \in \mathbb{R}^{d \times f}$ as its stacked stalk representation. At layer $l$, restriction maps are inferred via
\[
\mathcal{F}_{v \trianglelefteq e}^{(l)} =
\operatorname{reshape}_{d \times d}\!\left(
\sigma\big(
V^{(l)}(\operatorname{vec}(\mathbf{X}_v^{(l)}) \,\|\, \operatorname{vec}(\mathbf{X}_u^{(l)}))
\big)
\right),
\]
where $V^{(l)} \in \mathbb{R}^{d^2 \times 2df}$ is learnable and $\sigma$ is a nonlinearity. The nonlinear diffusion step is then
\begin{equation}
\mathbf{X}^{(l+1)} =
(1+\epsilon^{(l)})\mathbf{X}^{(l)} - \sigma\left(
\Delta_\mathcal{F}^{(l)} \left(I_{n} \otimes W_1^{(l)}\right)
\mathbf{X}^{(l)}
W_2^{(l)}\right) ,
\label{eq:nsd_update}
\end{equation}
where $W_1^{(l)} \in \mathbb{R}^{d \times d}$ and $W_2^{(l)} \in \mathbb{R}^{f \times f}$ act on stalk and feature dimensions, respectively. Restriction maps may be \emph{full}, \emph{orthogonal}, or \emph{diagonal}, trading off expressivity and parameter efficiency.

\paragraph{Related work.}
Neural Sheaf Diffusion was introduced by~\citet{bodnar2022neural} as a generalization of message passing using edge-conditioned linear maps. Subsequent works have explored extensions including precomputed maps~\citep{barbero2022sheaf}, complex-valued edge maps~\citep{amadocomplex}, attention-based variants~\citep{barbero2022sheafattention}, nonlinear sheaf diffusion~\citep{zaghen2024sheaf}, alternative communication schemes~\citep{ribeiro2025cooperative}, and multi-hop aggregation~\citep{bamberger2024bundle, borgi2025polynomial}. However, these approaches do not address the challenge of scaling sheaf-based architectures in depth.

\section{Deep Neural Sheaf Diffusion Architecture}
\label{sec:method}

Our goal is to design an architecture that maintains informative representations as the depth of aggregation increases. 

To this end, we introduce \textsc{DNSD}, a modified sheaf diffusion architecture. The key modification is the replacement of the sheaf Laplacian with a sheaf adjacency operator, which prevents the vanishing of diffusion signals at depth. We complement this change with additional design elements — normalization, odd nonlinearities, and gating — that improve stability under repeated feature mixing.  Detailed architectural specifications are provided in Appendix~\ref{app:architecture}.

We now describe the four modifications to the NSD update rule~\eqref{eq:nsd_update}. The modified update (DNSD) for Deep Neural Sheaf Diffusion takes the form:
\begin{equation}
\begin{aligned}
\mathbf{X}^{(l+1)} & = 
{\color{blue}\mathrm{LN}}\!\Big(
(1 + \epsilon^{(l)}) \mathbf{X}^{(l)} \\
&- {\color{blue}(\mathbf{G}^{(l)} \otimes \mathbf{1}_f^\top)} \odot 
\sigma_{\color{blue}\mathrm{odd}}\!\Big(
{\color{blue}A}_{\mathcal{F}}^{(l)} \left(I_{n} \otimes W_1^{(l)}\right)
\mathbf{X}^{(l)}
W_2^{(l)}
\Big)
\end{aligned}
\label{eq:sharp}
\end{equation}
where $A_\mathcal{F}$ is the sheaf adjacency operator, $\sigma_{\mathrm{odd}}$ is an odd activation, $\mathrm{LN}$ denotes layer normalization applied independently to each node stalk (i.e., row-wise over $\mathbf{X}_v^{(l)} \in \mathbb{R}^{d \times f}$), and $\mathbf{G}^{(l)} \in [0,1]^{nd}$ is the output of node-level per-stalk gate.

\paragraph{(i) Sheaf Convolution}
\label{sec:adjacency}

The original NSD~\eqref{eq:nsd_update} applies the nonlinearity to $\Delta_\mathcal{F} \mathbf{X}$, i.e. the sheaf Laplacian of the current signal.
In this setting, each diffusion step corresponds to a network layer, so increasing depth corresponds to iterating the diffusion process. Since $\Delta_\mathcal{F}$ measures disagreement between adjacent stalks, its output decreases as diffusion reduces this disagreement, eventually vanishing in the diffusion limit, i.e. at infinite depth~\citep{bodnar2022neural}. As a result, deeper layers tend to operate on progressively smaller residual signals and receive diminishing inputs to their nonlinearities.

This behavior reflects a limitation of using Laplacian-based updates in deep architectures. As depth increases, the model relies on $\Delta_\mathcal{F} \mathbf{X}$ to produce updates, yet this quantity is driven toward small magnitudes by the diffusion process itself. While this effect is formally established for linear diffusion, it remains a useful approximation of the behavior of neural sheaf diffusion in practice. Consequently, deeper layers can produce weaker updates and the loss can become less sensitive to their parameters. At initialization, a related issue appears: randomly initialized restriction maps can produce small and noisy disagreement signals, further reducing the effectiveness of deeper layers.

To address this limitation, we suggest to replace $\Delta_\mathcal{F}$ with $A_\mathcal{F}$, the sheaf adjacency operator, defined as block matrix with blocks $(A_\mathcal{F})_{uv} = \mathcal{F}_{u \trianglelefteq e}^\top \mathcal{F}_{v \trianglelefteq e}$ for $(u,v) \in \mathcal{E}$. This yields a sheaf convolution, where updates are computed by aggregating dependency between neighboring representations rather than measuring residual disagreement. The resulting update $\sigma(A_\mathcal{F} \left(I_{n} \otimes W_1^{(l)}\right)
\mathbf{X}^{(l)}
W_2^{(l)})$ therefore operates on an informative signal, ensuring meaningful inputs to the nonlinearity both at initialization and throughout depth.

\begin{remark}[Relation to prior formulations]
\citet{bodnar2022neural} introduced this sheaf convolution operator as part of a discretization of sheaf diffusion (their Eq.~4). However, their final architecture applies the nonlinearity only to the Laplacian term, effectively removing the identity component and feeding only the disagreement signal to the activation. As discussed above, this choice drives the signal toward small magnitudes at depth, while retaining the full adjacency operator preserves informative signals throughout the network.
\end{remark}

\paragraph{(ii) Layer Normalization}
\label{sec:layernorm}

Replacing the Laplacian with the adjacency operator ensures that updates remain informative at depth. However, repeated application of these updates still introduces a second challenge: as depth increases, the scale of node representations can vary significantly across layers due to accumulated transformations and nonlinearities. This variation makes optimization unstable, as different layers operate on signals with inconsistent magnitudes.

To address this issue, layer normalization~\citep{ba2016layer} is applied to the representation $\tilde{\mathbf{X}}_u^{(l)} \in \mathbb{R}^{d \times f}$ obtained \emph{after} aggregation and residual addition, independently for each stalk by normalizing across the feature dimension $f$ as
\begin{equation}
\mathrm{LN}^{(l)}(\tilde{\mathbf{X}}_u^{(l)})
= \mathbf{1}_d \gamma^{(l)\top} \odot
\frac{\tilde{\mathbf{X}}_u^{(l)} - \mu_u \mathbf{1}_f^\top}
{\sigma_u \mathbf{1}_f^\top}
+ \mathbf{1}_d \beta^{(l)\top}
\label{eq:layernorm}
\end{equation}
where $\mu_u, \sigma_u \in \mathbb{R}^{d}$ are the per-stalk mean and standard deviation computed across the feature dimension $f$, and $\gamma^{(l)}, \beta^{(l)} \in \mathbb{R}^{f}$ are learnable parameters shared across all nodes and stalks. By normalizing each representation at every layer, LayerNorm stabilizes both the forward signal and the backward gradient, ensuring consistent magnitudes throughout depth. The learnable affine parameters $(\gamma^{(l)}, \beta^{(l)})$ allow the model to adapt the scale and offset of normalized features at each layer.

\paragraph{(iii) Odd and Bounded Activation Functions}
\label{sec:odd}

While the adjacency formulation preserves an informative signal at depth, deep architectures repeatedly apply the same transformation across layers. In this regime, the choice of nonlinearity directly affects how representations evolve under repeated composition.

The NSD in \eqref{eq:nsd_update} uses ReLU as its pointwise nonlinearity. ReLU introduces an asymmetric truncation: negative activations are suppressed, while positive ones are preserved. In our setting, where the message term is subtracted from a residual, this results in a one-sided update that can only decrease coordinates. As depth increases, this asymmetric suppression accumulates, progressively reducing variation across features and inducing drift in the representation geometry.

To address this issue, we replace ReLU with an odd and bounded activation function $\sigma_{\mathrm{odd}}$ (specifically $\tanh$). Odd activations preserve sign symmetry, allowing both positive and negative interactions to contribute equally, while boundedness controls the magnitude of updates under repeated composition. Together, these properties maintain a balanced and stable transformation of representations across layers, which is particularly important when stacking many layers.

\paragraph{(iv) Node-Level Per-Stalk Gating}
\label{sec:gating}
While the previous modifications ensure that signals remain informative and transformations remain stable at depth, repeated weighted aggregation introduces an additional challenge: each layer combines contributions from neighboring nodes and stalk dimensions, and these contributions may contain noise. As depth increases, such noisy components can accumulate under repeated composition, reducing the quality of representations at increasing depth.

This phenomenon affects NSD, which applies updates uniformly across all representation dimensions as in ~\eqref{eq:nsd_update}. However, not all contributions are equally relevant at every node or layer. Under repeated composition, this uniform treatment propagates all components indiscriminately, allowing noisy contributions to persist and accumulate. This phenomenon is analogous to the \emph{attention sink} effect observed in deep attention models~\citep{xiao2023efficient, darcet2023vision}, where certain components accumulate disproportionate signal without contributing meaningfully to the representation, thereby degrading its quality.

Therefore, we suggest to introduce a node-level gating mechanism that selectively modulates the contribution of each representation dimension at every layer, that is,
\begin{equation}
\left[(\mathbf{G}^{(l)})_u\right]_s = \mathrm{sigmoid}\!\left(w_g^{(l)} \begin{bmatrix}\mathbf{X}_{u,s}^{(l)} \\ \bar{\mathbf{X}}_{u,s}^{(l)}\end{bmatrix} + b_g^{(l)}\right) \in [0,1]
\label{eq:gate}
\end{equation}
where $w_g^{(l)} \in \mathbb{R}^{1 \times 2f}$ and $b_g^{(l)} \in \mathbb{R}$ are shared across all stalks, and $\mathbf{X}_{u,s}^{(l)}, \bar{\mathbf{X}}_{u,s}^{(l)} \in \mathbb{R}^f$ denote the $s$-th stalk of the current embedding and of the aggregated signal before $\sigma$, respectively.
The gate modulates the update in~\eqref{eq:sharp} via $(\mathbf{G}^{(l)} \otimes \mathbf{1}_f^\top) \odot (\cdot)$, broadcasting each per-stalk scalar uniformly over the $f$ feature dimensions.

This selective filtering limits the accumulation of noisy components under repeated composition, preserving representation quality when stacking layers.

\begin{table*}[!t]
\centering
\caption{Comparison of message passing mechanisms along key design axes: the update operator (difference- vs dependency-based), the form of edge transformations (scalar vs matrix-valued), the normalization strategy, and the resulting behavior at depth.} 
\label{tab:sheaf_attention}
\resizebox{\textwidth}{!}{%
\begin{tabular}{lcccc}
\toprule
\textbf{Model} 
& \textbf{Update operator} 
& \textbf{Edge transformation} 
& \textbf{Normalization} 
& \textbf{Behavior at depth} \\
\midrule
GATs~\citep{ brody2021attentive}
& dependency (adjacency operator)
& scalar (attention scores) 
& attention scores (softmax) 
& averaging tendency \\

\midrule
NSD~\citep{bodnar2022neural}
& difference (laplacian)  
& linear map (matrix-valued) 
& operator (degree normalization)
& vanishing signal \\

\textsc{DNSD} (ours)
& dependency (adjacency operator)
& linear map (matrix-valued) 
& operator + representations (LN)
& mitigates signal decay \\
\bottomrule
\end{tabular}%
}
\end{table*}%

\section{Sheaf Diffusion through the Lens of Graph Attention}
\label{sec:attention}

We analyze \textsc{DNSD} through the lens of graph attention mechanisms, focusing on how it relates to existing message passing architectures. We use “GATs” to refer to the class of graph attention models analyzed in~\citet{brody2021attentive}.  

In \textsc{DNSD} the residual update at layer $l$ at node $u$ can be written as
\(
\sum_{v \in \mathcal{N}(u)} [\mathbf{A}_{\mathcal{F}}^{(l)}]_{uv} \mathbf{X}_v^{(l)},
\)
where $[\mathbf{A}_{\mathcal{F}}^{(l)}]_{uv}$ are edge operators computed by a learnable function of pairwise node representations $(\mathbf{X}_u^{(l)}, \mathbf{X}_v^{(l)})$. This update aggregates neighboring node representations using edge-dependent functions of node pairs, aligning it with attention mechanisms where aggregation weights are likewise computed from such pairwise interactions. However, DNSD differs from standard graph attention along several structural axes, as summarized in Table~\ref{tab:sheaf_attention}.

\paragraph{Update operator.}
Graph attention mechanisms and DNSD both aggregate \emph{dependency} between neighboring node representations through adjacency-based operators. In contrast, NSD applies the Laplacian, which measures \emph{difference} between neighboring node representations.
While DNSD and attention share this dependency-based formulation, they differ in how these interactions are represented and combined, as discussed next.

\paragraph{Edge functions.}
Graph attention mechanisms and DNSD both compute edge-dependent functions from pairs of node representations. In attention, these functions map $(\mathbf{X}_u^{(l)}, \mathbf{X}_v^{(l)})$ to scalar attention scores that weight neighboring representations. In contrast, \textsc{DNSD} maps these pairs to linear maps $\mathcal{F}_{uv}^{(l)}$, which are applied directly to node representations.

\paragraph{Normalization.}
Attention mechanisms normalize scalar attention scores using softmax, which enforces non-negative weights that sum to one and results in convex aggregation of neighboring node representations. In contrast, \textsc{DNSD} does not normalize the aggregation operators, but instead applies normalization to node representations after aggregation. While DNSD inherits operator normalization from the sheaf construction, the dominant normalization occurs at the level of node representations, allowing aggregation to be non-convex.

\paragraph{Behavior at depth.}
These mechanisms lead to distinct behavior at increasing depth. In attention-based models, softmax normalization of scalar scores induces convex aggregation, resulting in an averaging tendency and potential representation collapse~\citep{wu2023demystifying}. In NSD, updates are computed from differences between neighboring representations, which decrease as diffusion progresses, leading to vanishing signals. DNSD mitigates both effects by combining dependency-based aggregation with representation-level normalization, maintaining informative signals across layers.

\section{Experiments}
\label{sec:experiments}

In this section, 
we present the results of two sets of experiments, one with a synthetic long-range dataset, and one with real-world heterophilic benchmarks.
\begin{figure*}[!tbp]
    \centering
    \includegraphics[width=0.9\linewidth]{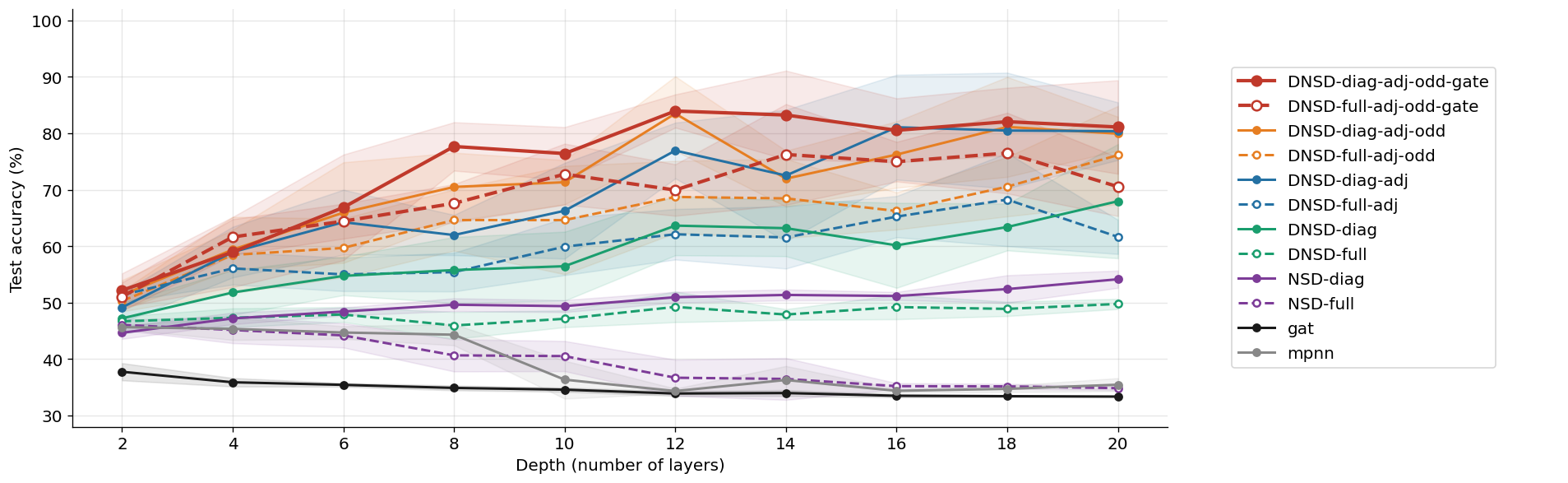}
    \caption{\textbf{Test accuracy on the synthetic community detection dataset G5 as a function of depth.} Each curve corresponds to a model variant, where \texttt{adj}, \texttt{odd}, and \texttt{gate} denote the use of adjacency-based operator, odd nonlinearities, and gating, respectively, and \texttt{diag} / \texttt{full} refer to diagonal or full restriction maps. Shaded regions represent one standard deviation over multiple runs.}
    \label{fig:g5_depth}
\end{figure*}

\paragraph{Synthetic Experiments.} We design a dataset with provable long-lange dependence, suitable benchmarking deep MPNNs as follows.
Building on the synthetic experiments from \citep{zaghen2024sheaf}, we construct different sets of graphs based on the following process. We consider a family graph with 3 communities of 500 nodes each, with node features as 2D Gaussian features ($\sigma=3$, shifted centers respectively $(0,0), (1,0), (0,1)$). The connectivity is defined by connecting each node to k-nearest neighbours ($k=8$) within its community, forming three homophilic KNN graphs. We then gradually introduce heterophilic noise, in each step rewiring a fraction ($10\%$) of the initial  intra-community  edges to random cross-community nodes. This leads to eleven perturbation levels from G0 (homophilic) to G10 (fully heterophilic). Solving this problem requires a deep receptive field of the GNN. On the one hand, guessing the right label must be performed by aggregating a sufficient number of features to compute the mean of community features. On the other hand, possible combinatorial ties have to be resolved for entire connected components. 
In the experiments training uses 6 graphs (9000 nodes, seeds 42--47) with 80/20 train/val split, testing uses 3 separate graphs (4500 nodes, seeds 100--102). Learning is evaluated based on the 3-class node classification accuracy on all test nodes.

We compare \textit{\textsc{DNSD}} variants by gradually adding (i) adjacency, (iii) odd activation, and (iv) gating. For all \textit{\textsc{DNSD}} variants, normalization is applied, as this is standard practice when training deep neural networks. We compare these models, using both full and diagonal maps, against \textit{NSD} from \citep{bodnar2022neural} and the non-sheaf baselines MLP, MPNN~\citep{gilmer2017neural}, and GAT~\citep{velivckovic2017graph}. We restrict our experiments to diagonal and full restriction maps; orthogonal maps were found to be difficult to train robustly at depth in our setting, and we leave a systematic study of orthogonal parameterizations to future work. To assess the effect of depth, we evaluate all models across a range of layers $\{2,4,8,12,16\}$. Experimental details are given in Appendix~\ref{app:experiments_synthetic}.

\begin{table*}[t]
\centering
\caption{Synthetic community detection: test accuracy (\%) across perturbation levels G0--G10 on the test sets. All \textsc{DNSD} variants include LayerNorm. Each entry reports mean\,$\pm$\,std at the best layer depth (in parentheses) for that (level, model) pair. \rone{1st}, \rtwo{2nd}, \rthree{3rd} best per column. }
\label{tab:synthetic}
\resizebox{\textwidth}{!}{%
\begin{tabular}{@{}l l ccc ccccccccccc@{}}
\toprule
& & \multicolumn{3}{c}{\textbf{Fixes}} & \multicolumn{11}{c}{\textbf{Perturbation level}} \\
\cmidrule(lr){3-5} \cmidrule(lr){6-16}
& \textbf{Map} & \rotatebox{60}{Adj} & \rotatebox{60}{Odd} & \rotatebox{60}{Gate}
& G0 & G1 & G2 & G3 & G4 & G5 & G6 & G7 & G8 & G9 & G10 \\
\midrule
MLP  & --- & & &
& 41.3\sd{0.3}{\scriptsize(L2)} & 41.4\sd{0.2}{\scriptsize(L2)} & 41.3\sd{0.3}{\scriptsize(L2)} & 41.3\sd{0.3}{\scriptsize(L2)} & 41.3\sd{0.1}{\scriptsize(L2)} & 41.3\sd{0.2}{\scriptsize(L2)} & 41.2\sd{0.2}{\scriptsize(L2)} & 41.2\sd{0.3}{\scriptsize(L2)} & 41.3\sd{0.2}{\scriptsize(L2)} & 41.2\sd{0.2}{\scriptsize(L2)} & 41.5\sd{0.2}{\scriptsize(L2)} \\
MPNN & --- & & &
& 53.9\sd{2.0}{\scriptsize(L2)} & 47.0\sd{1.0}{\scriptsize(L2)} & 49.6\sd{3.3}{\scriptsize(L4)} & 47.7\sd{1.2}{\scriptsize(L2)} & 47.1\sd{1.5}{\scriptsize(L2)} & 45.6\sd{0.4}{\scriptsize(L2)} & 45.0\sd{1.1}{\scriptsize(L2)} & 45.6\sd{0.7}{\scriptsize(L2)} & 46.6\sd{0.5}{\scriptsize(L2)} & 50.2\sd{1.0}{\scriptsize(L4)} & 94.7\sd{2.2}{\scriptsize(L8)} \\
GAT  & --- & & &
& \rone{67.2\sd{15.9}{\scriptsize(L8)}} & 45.6\sd{2.1}{\scriptsize(L4)} & 47.9\sd{2.0}{\scriptsize(L4)} & 43.9\sd{1.2}{\scriptsize(L4)} & 38.6\sd{2.2}{\scriptsize(L4)} & 37.7\sd{1.5}{\scriptsize(L2)} & 36.7\sd{1.8}{\scriptsize(L2)} & 36.1\sd{1.0}{\scriptsize(L2)} & 41.8\sd{1.1}{\scriptsize(L2)} & 46.7\sd{1.1}{\scriptsize(L4)} & 64.8\sd{2.0}{\scriptsize(L4)} \\
\midrule
\multirow{2}{*}{NSD}
& diag & & &
& 53.0\sd{16.4}{\scriptsize(L12)} & 51.0\sd{6.4}{\scriptsize(L16)} & 55.8\sd{0.9}{\scriptsize(L12)} & 54.7\sd{2.3}{\scriptsize(L16)} & 51.2\sd{2.1}{\scriptsize(L16)} & 51.2\sd{0.7}{\scriptsize(L16)} & 49.1\sd{1.7}{\scriptsize(L16)} & 49.1\sd{1.2}{\scriptsize(L12)} & 50.1\sd{1.0}{\scriptsize(L16)} & 50.5\sd{1.5}{\scriptsize(L16)} & 85.5\sd{4.7}{\scriptsize(L16)} \\
& full & & &
& 53.1\sd{5.6}{\scriptsize(L16)} & 46.4\sd{1.4}{\scriptsize(L2)} & 49.2\sd{1.8}{\scriptsize(L4)} & 46.0\sd{0.5}{\scriptsize(L2)} & 46.4\sd{1.2}{\scriptsize(L2)} & 46.1\sd{0.9}{\scriptsize(L2)} & 45.2\sd{0.7}{\scriptsize(L2)} & 45.5\sd{0.4}{\scriptsize(L2)} & 46.7\sd{0.7}{\scriptsize(L2)} & 49.3\sd{1.1}{\scriptsize(L8)} & 84.0\sd{4.0}{\scriptsize(L12)} \\
\midrule
\multirow{5}{*}{\rotatebox{90}{\textsc{DNSD} (diag)}}
& diag & \cmark & \cmark & \cmark
& 45.7\sd{10.1}{\scriptsize(L16)} & 51.0\sd{9.1}{\scriptsize(L4)} & \rtwo{73.9\sd{5.7}{\scriptsize(L12)}} & \rone{83.5\sd{4.3}{\scriptsize(L12)}} & \rone{82.4\sd{7.3}{\scriptsize(L12)}} & \rone{83.9\sd{2.9}{\scriptsize(L12)}} & \rone{86.1\sd{1.8}{\scriptsize(L12)}} & \rtwo{81.5\sd{5.5}{\scriptsize(L16)}} & \rtwo{75.6\sd{4.7}{\scriptsize(L16)}} & \rtwo{63.4\sd{4.4}{\scriptsize(L12)}} & 96.2\sd{1.3}{\scriptsize(L16)} \\
& diag & \cmark & \cmark &
& 51.1\sd{6.1}{\scriptsize(L16)} & \rone{57.3\sd{3.5}{\scriptsize(L12)}} & \rone{74.5\sd{1.9}{\scriptsize(L12)}} & \rtwo{75.4\sd{11.1}{\scriptsize(L12)}} & \rtwo{79.6\sd{8.7}{\scriptsize(L12)}} & \rtwo{83.5\sd{6.6}{\scriptsize(L12)}} & \rtwo{83.3\sd{4.5}{\scriptsize(L12)}} & \rone{82.2\sd{4.5}{\scriptsize(L16)}} & \rone{77.1\sd{2.0}{\scriptsize(L16)}} & \rone{64.3\sd{1.8}{\scriptsize(L16)}} & 96.7\sd{0.7}{\scriptsize(L16)} \\
& diag & \cmark &  & \cmark
& 48.0\sd{9.5}{\scriptsize(L4)} & 49.9\sd{7.6}{\scriptsize(L16)} & 58.2\sd{1.8}{\scriptsize(L16)} & 62.0\sd{10.1}{\scriptsize(L8)} & \rthree{72.5\sd{6.2}{\scriptsize(L12)}} & \rthree{80.4\sd{9.4}{\scriptsize(L12)}} & \rthree{81.0\sd{8.0}{\scriptsize(L12)}} & \rthree{77.3\sd{5.2}{\scriptsize(L12)}} & \rthree{74.4\sd{6.3}{\scriptsize(L12)}} & \rthree{61.0\sd{3.3}{\scriptsize(L12)}} & 95.7\sd{1.4}{\scriptsize(L16)} \\
& diag &  & \cmark & \cmark
& \rtwo{59.2\sd{20.9}{\scriptsize(L12)}} & 51.0\sd{13.2}{\scriptsize(L16)} & 63.3\sd{6.2}{\scriptsize(L16)} & \rthree{65.5\sd{14.9}{\scriptsize(L16)}} & 66.9\sd{8.0}{\scriptsize(L12)} & 71.4\sd{11.8}{\scriptsize(L16)} & 71.1\sd{7.5}{\scriptsize(L16)} & 70.6\sd{7.4}{\scriptsize(L16)} & 62.8\sd{5.2}{\scriptsize(L16)} & 55.5\sd{4.3}{\scriptsize(L16)} & 95.3\sd{2.6}{\scriptsize(L16)} \\
& diag &  &  &
& 50.4\sd{6.4}{\scriptsize(L16)} & 49.8\sd{5.0}{\scriptsize(L16)} & 54.9\sd{4.1}{\scriptsize(L4)} & 53.5\sd{5.3}{\scriptsize(L16)} & 60.8\sd{9.8}{\scriptsize(L16)} & 63.6\sd{5.3}{\scriptsize(L12)} & 60.4\sd{7.7}{\scriptsize(L16)} & 58.8\sd{6.2}{\scriptsize(L16)} & 57.8\sd{4.5}{\scriptsize(L16)} & 53.4\sd{1.7}{\scriptsize(L16)} & 96.4\sd{0.8}{\scriptsize(L16)} \\
\midrule
\multirow{5}{*}{\rotatebox{90}{\textsc{DNSD} (full)}}
& full & \cmark & \cmark & \cmark
& 52.5\sd{4.4}{\scriptsize(L12)} & 52.5\sd{4.8}{\scriptsize(L12)} & 65.4\sd{3.3}{\scriptsize(L12)} & 63.3\sd{4.9}{\scriptsize(L16)} & 69.0\sd{3.7}{\scriptsize(L16)} & 75.0\sd{3.5}{\scriptsize(L16)} & 72.4\sd{5.1}{\scriptsize(L16)} & 69.0\sd{2.6}{\scriptsize(L16)} & 64.6\sd{4.4}{\scriptsize(L16)} & 56.1\sd{1.9}{\scriptsize(L12)} & \rthree{97.1\sd{0.7}{\scriptsize(L16)}} \\
& full & \cmark & \cmark &
& 52.1\sd{2.0}{\scriptsize(L12)} & \rtwo{54.0\sd{3.9}{\scriptsize(L16)}} & \rthree{64.3\sd{4.4}{\scriptsize(L16)}} & 61.2\sd{3.2}{\scriptsize(L12)} & 63.0\sd{3.4}{\scriptsize(L16)} & 68.7\sd{6.4}{\scriptsize(L12)} & 63.5\sd{5.8}{\scriptsize(L16)} & 59.5\sd{2.3}{\scriptsize(L12)} & 55.3\sd{2.8}{\scriptsize(L12)} & 53.2\sd{1.3}{\scriptsize(L16)} & \rtwo{97.3\sd{1.3}{\scriptsize(L16)}} \\
& full & \cmark &  & \cmark
& 51.7\sd{2.3}{\scriptsize(L16)} & \rthree{53.3\sd{4.8}{\scriptsize(L8)}} & 61.7\sd{4.9}{\scriptsize(L8)} & 61.7\sd{5.4}{\scriptsize(L12)} & 68.0\sd{11.0}{\scriptsize(L16)} & 73.0\sd{4.7}{\scriptsize(L16)} & 66.2\sd{7.7}{\scriptsize(L16)} & 63.9\sd{5.3}{\scriptsize(L16)} & 63.3\sd{5.0}{\scriptsize(L16)} & 55.2\sd{1.6}{\scriptsize(L16)} & \rone{97.5\sd{0.8}{\scriptsize(L16)}} \\
& full &  & \cmark & \cmark
& \rthree{54.3\sd{5.6}{\scriptsize(L16)}} & 51.4\sd{4.4}{\scriptsize(L16)} & 58.9\sd{1.7}{\scriptsize(L16)} & 56.6\sd{2.7}{\scriptsize(L16)} & 57.2\sd{4.3}{\scriptsize(L16)} & 58.0\sd{2.6}{\scriptsize(L12)} & 54.3\sd{5.5}{\scriptsize(L12)} & 53.7\sd{3.2}{\scriptsize(L16)} & 53.1\sd{2.7}{\scriptsize(L16)} & 52.7\sd{1.0}{\scriptsize(L12)} & 95.5\sd{1.6}{\scriptsize(L16)} \\
& full &  &  &
& 47.5\sd{0.8}{\scriptsize(L12)} & 47.9\sd{4.1}{\scriptsize(L2)} & 52.4\sd{1.6}{\scriptsize(L16)} & 49.0\sd{2.4}{\scriptsize(L16)} & 50.3\sd{3.2}{\scriptsize(L16)} & 49.2\sd{2.7}{\scriptsize(L12)} & 46.6\sd{2.5}{\scriptsize(L16)} & 46.4\sd{1.8}{\scriptsize(L16)} & 47.6\sd{1.1}{\scriptsize(L16)} & 50.4\sd{0.9}{\scriptsize(L16)} & 96.7\sd{1.9}{\scriptsize(L16)} \\
\bottomrule
\end{tabular}%
}
\end{table*}

\paragraph{Real-World Experiments}
We evaluate on the heterophilic benchmark suite of~\citet{platonov2023critical} as a standard real-world setting, including large-scale graphs such as Penn94~\citep{lim2021large}. Following the concerns raised by~\citet{bechler2025position} regarding the ability of current benchmarks to isolate specific modeling factors, we use the synthetic benchmark to explicitly study depth, while the real-world benchmark serve as a complementary evaluation of robustness across diverse graph structures.
We re-run NSD and \textsc{DNSD} under the same hyperparameter budget. Due to computational constraints, we restrict the number of layers to at most 8 and focus on diagonal maps, as motivated by results on synthetic benchmark in Table~\ref{tab:synthetic}, which offer a favorable trade-off between performance and efficiency.
Details can be found in Appendix~\ref{app:real_world_experiments}.

\begin{table*}[tpb]
\centering
\caption{Node classification accuracy (\%) on heterophilic benchmarks. All \textsc{DNSD} variants use diagonal restriction maps and include LayerNorm. Each entry reports mean\,$\pm$\,std at the best layer depth (in parentheses) for that (dataset, model) pair. \rone{1st}, \rtwo{2nd}, \rthree{3rd} best per column.}
\label{tab:realworld}
\resizebox{0.7\textwidth}{!}{%
\begin{tabular}{@{}l l ccc cccccc@{}}
\toprule
& & \multicolumn{3}{c}{\textbf{Fixes}} & \multicolumn{6}{c}{\textbf{Dataset}} \\
\cmidrule(lr){3-5} \cmidrule(lr){6-11}
& \textbf{Map}
& \rotatebox{0}{Adj} & \rotatebox{0}{Odd} & \rotatebox{0}{Gate}
& \rotatebox{0}{\makecell{Roman\\Empire}}
& \rotatebox{0}{\makecell{Amazon\\Ratings}}
& \rotatebox{0}{Minesweeper}
& \rotatebox{0}{Tolokers}
& \rotatebox{0}{Questions}
& \rotatebox{0}{Penn94} \\
\midrule
MLP  & --- & & & & 66.4\sd{0.1}{\scriptsize(L2)} & 40.9\sd{0.4}{\scriptsize(L2)} & 80.0\sd{0.0}{\scriptsize(L2)} & 78.2\sd{0.0}{\scriptsize(L2)} & 97.0\sd{0.0}{\scriptsize(L2)} & 76.2\sd{0.0}{\scriptsize(L2)} \\
MPNN & --- & & & &
 78.9\sd{0.8}{\scriptsize(L2)} & 46.6\sd{0.3}{\scriptsize(L4)} & 87.4\sd{1.3}{\scriptsize(L4)} & 79.1\sd{0.1}{\scriptsize(L2)} & 97.0\sd{0.0}{\scriptsize(L2)} & --- \\
GAT  & --- & & & &
 56.9\sd{1.1}{\scriptsize(L2)} & 46.0\sd{0.6}{\scriptsize(L2)} & 80.3\sd{0.0}{\scriptsize(L2)} & 78.4\sd{0.1}{\scriptsize(L4)} & 97.0\sd{0.0}{\scriptsize(L2)} & 74.1\sd{2.0}{\scriptsize(L2)} \\
\midrule
\multirow{2}{*}{NSD}
& diag & & & &
 79.1\sd{0.5}{\scriptsize(L8)} & 44.6\sd{0.3}{\scriptsize(L8)} & 87.5\sd{0.6}{\scriptsize(L8)} & 81.5\sd{0.2}{\scriptsize(L8)} & 97.1\sd{0.0}{\scriptsize(L2)} & 76.3\sd{0.1}{\scriptsize(L2)} \\
& full & & & &
 77.1\sd{0.9}{\scriptsize(L4)} & 45.4\sd{0.7}{\scriptsize(L4)} & 86.1\sd{0.3}{\scriptsize(L4)} & 81.4\sd{0.3}{\scriptsize(L2)} & 97.1\sd{0.0}{\scriptsize(L2)} & 76.1\sd{0.6}{\scriptsize(L4)} \\
\midrule
\multirow{5}{*}{\rotatebox{90}{\textsc{DNSD} (diag)}}
& diag & \cmark & \cmark & \cmark &
 \rtwo{83.4\sd{0.9}{\scriptsize(L8)}} & \rtwo{47.8\sd{0.4}{\scriptsize(L8)}} & \rone{89.4\sd{0.8}{\scriptsize(L8)}} & \rtwo{81.8\sd{0.5}{\scriptsize(L4)}} & 97.1\sd{0.0}{\scriptsize(L4)} & \rtwo{78.7\sd{0.9}{\scriptsize(L8)}} \\
& diag & \cmark & \cmark &  &
 \rthree{83.2\sd{0.7}{\scriptsize(L8)}} & \rone{49.1\sd{0.7}{\scriptsize(L8)}} & \rtwo{88.9\sd{0.4}{\scriptsize(L8)}} & \rone{82.0\sd{0.2}{\scriptsize(L8)}} & 97.1\sd{0.0}{\scriptsize(L2)} & \rone{80.0\sd{0.9}{\scriptsize(L8)}} \\
& diag & \cmark &  & \cmark &
 83.0\sd{0.4}{\scriptsize(L8)} & \rthree{47.5\sd{1.0}{\scriptsize(L4)}} & 88.1\sd{0.8}{\scriptsize(L8)} & \rthree{81.8\sd{0.7}{\scriptsize(L8)}} & 97.1\sd{0.0}{\scriptsize(L4)} & \rthree{78.6\sd{0.9}{\scriptsize(L8)}} \\
& diag &  & \cmark & \cmark &
  \rone{83.4\sd{0.2}{\scriptsize(L8)}} & 46.0\sd{0.4}{\scriptsize(L2)} & \rthree{88.2\sd{0.5}{\scriptsize(L8)}} & 81.5\sd{0.4}{\scriptsize(L8)} & \rone{97.2\sd{0.0}{\scriptsize(L2)}} & 76.0\sd{0.6}{\scriptsize(L4)} \\
& diag &  &  &  &
 82.7\sd{0.5}{\scriptsize(L8)} & 46.8\sd{0.5}{\scriptsize(L8)} & 87.6\sd{0.6}{\scriptsize(L8)} & 81.1\sd{0.4}{\scriptsize(L2)} & \rtwo{97.2\sd{0.0}{\scriptsize(L8)}} & 76.5\sd{0.4}{\scriptsize(L8)} \\
\bottomrule
\end{tabular}%
}
\end{table*}

\paragraph{Results on synthetic experiments.}
Table~\ref{tab:synthetic} reports test accuracy across perturbation levels together with the depth of the best-performing models. Baselines perform poorly across all settings, often close to random guessing.
For \textsc{DNSD}, the adjacency operator is the dominant factor: variants with adjacency outperform NSD by 20--30 percentage points at mid-heterophily levels (G3--G9), while variants without adjacency remain near the baseline. The odd activation provides additional gains, particularly at moderate heterophily (G2--G5), whereas gating mainly reduces variance across runs. Overall, the best performance is achieved by configurations combining adjacency, odd activation, and gating, followed by variants without gating, and then without both gating and odd activation.

\paragraph{Depth behavior.}
Figure~\ref{fig:g5_depth} shows performance as a function of depth. \textsc{DNSD} improves with depth, reaching its best performance between 12 and 16 layers, consistent with Table~\ref{tab:synthetic}. Similar trends are observed on real-world benchmarks (Table~\ref{tab:realworld}), where \textsc{DNSD} typically achieves its best performance at maximum depths 8. In contrast, graph attention models degrade at larger depths, while NSD shows limited improvement and often plateaus or decreases. \textsc{DNSD} remains effective at larger depths, unlike these baselines.

\paragraph{Results on real-world datasets.}
On real-world heterophilic benchmarks (Table~\ref{tab:realworld}), gains are smaller but remain consistent across datasets. \textsc{DNSD} (diag, Adj+Odd) achieves the best or second-best performance on 5 out of 9 datasets, with the largest improvements on Roman Empire (+4.1pp) and Penn94 (+3.7pp over NSD diag).
Across datasets, diagonal restriction maps consistently match or outperform full maps, despite having fewer parameters.

\section{Discussion Future Directions and Conclusions}
\label{sec:conclusion}

\paragraph{Discussion of architectural choices.}
The experimental results highlight a clear hierarchy among the proposed modifications. The dominant factor is the replacement of the Laplacian with the adjacency operator, which shifts the update from measuring difference to aggregating dependency between neighboring representations. This change consistently yields the largest gains and is the primary driver of improved behavior at depth. The use of diagonal restriction maps further shows that that high-dimensional restriction maps are not required for strong performance, and that simpler per-stalk interactions suffice and are even beneficial in practice, consistent with~\citet{borgi2025polynomial}. Additional components like odd nonlinearities and gating mainly act as stabilizing mechanisms: odd activations prevent biased geometric shift of representations under repeated composition, while gating reduces variance across runs without significantly affecting peak performance.

\paragraph{Limitations and Future Directions.}
\textsc{DNSD} is evaluated on node classification tasks on medium-scale graphs. This limits the assessment of its applicability across diverse graph learning problems. Extending the evaluation to graph-level tasks, link prediction, and larger-scale datasets is an important direction, particularly in the context of graph foundation models~\citep{wang2025graph}.

Our analysis of depth in real-world settings is limited by the lack of benchmarks that effectively probe deep GNN behavior, reflecting broader concerns that current graph benchmarks do not isolate properties such as depth and long-range dependencies~\citep{bechler2025position}. While the observed trends are consistent with the synthetic benchmark, current datasets do not allow for a systematic evaluation of deeper architectures. Developing benchmarks that explicitly capture long-range dependencies remains an important direction for future work.

The sheaf-based formulation introduces richer edge interactions, which may increase computational cost at scale. This raises questions about scalability and efficiency. Investigating these trade-offs in distributed settings, where sheaves can model inter-agent agreement constraints, is a promising direction for future work~\citep{hanks2025distributed, bourgerie2026euclidean}.

\paragraph{Conclusion.}
We introduce \textsc{DNSD}, a sheaf diffusion architecture for deep GNNs. The key insight is that effective depth requires dependency-based propagation together with explicit normalization of representations across layers. By replacing Laplacian updates with adjacency-based aggregation, \textsc{DNSD} yields architectures that remain effective as depth increases, as demonstrated on both synthetic and real-world benchmarks. 
These results position \textsc{DNSD} as a promising architectural direction for graph foundation models requiring long range propagation.
\section*{Acknowledgement}
The computations were enabled by resources provided by the National Academic Infrastructure for Supercomputing in Sweden (NAISS), partially funded by the Swedish Research Council through grant agreement no. 2022-06725.
\balance
\bibliography{references}

@inproceedings{bodnar2022neural,
  title={Neural Sheaf Diffusion: A Topological Perspective on Heterophily and Oversmoothing in {GNN}s},
  author={Bodnar, Cristian and Di Giovanni, Francesco and Chamberlain, Benjamin P. and Li\`{o}, Pietro and Bronstein, Michael},
  booktitle={Advances in Neural Information Processing Systems},
  volume={35},
  pages={18527--18541},
  year={2022}
}

@inproceedings{gilmer2017neural,
  title={Neural Message Passing for Quantum Chemistry},
  author={Gilmer, Justin and Schoenholz, Samuel S and Riley, Patrick F and Vinyals, Oriol and Dahl, George E},
  booktitle={International Conference on Machine Learning},
  pages={1263--1272},
  year={2017}
}

@inproceedings{velivckovic2017graph,
  title={Graph Attention Networks},
  author={Veli{\v{c}}kovi{\'c}, Petar and Cucurull, Guillem and Casanova, Arantxa and Romero, Adriana and Li{\`o}, Pietro and Bengio, Yoshua},
  booktitle={International Conference on Learning Representations},
  year={2018}
}

@inproceedings{brody2021attentive,
  title={How Attentive Are Graph Attention Networks?},
  author={Brody, Shaked and Alon, Uri and Yahav, Eran},
  booktitle={International Conference on Learning Representations},
  year={2022}
}

@inproceedings{platonov2023critical,
  title={A Critical Look at the Evaluation of {GNN}s Under Heterophily: Are We Really Making Progress?},
  author={Platonov, Oleg and Kuznedelev, Denis and Diskin, Michael and Babenko, Artem and Prokhorenkova, Liudmila},
  booktitle={International Conference on Learning Representations},
  year={2023}
}

@inproceedings{barbero2022sheaf,
  title={Sheaf Neural Networks with Connection Laplacians},
  author={Barbero, Federico and Bodnar, Cristian and de Oc{\'a}riz Borde, Haitz S{\'a}ez and Bronstein, Michael and Veli{\v{c}}kovi{\'c}, Petar and Li{\`o}, Pietro},
  booktitle={Proceedings of Topological, Algebraic, and Geometric Learning Workshops 2022},
  pages={28--36},
  volume={196},
  organization={PMLR},
  year={2022}
}

@inproceedings{ribeiro2025cooperative,
  title={Cooperative Sheaf Neural Networks},
  author={Ribeiro, Andr{\'e} and Ten{\'o}rio, Ana Luiza and Belieni, Juan and Souza, Amauri H and Mesquita, Diego},
  booktitle={International Conference on Learning Representations},
  year={2026}
}

@inproceedings{qiu2025gated,
  title={Gated Attention for Large Language Models: Non-Linearity, Sparsity, and Attention-Sink-Free},
  author={Qiu, Zihan and Wang, Zekun and Zheng, Bo and Huang, Zeyu and Wen, Kaiyue and Yang, Songlin and Men, Rui and Yu, Le and Huang, Fei and Huang, Suozhi and Liu, Dayiheng and Zhou, Jingren and Lin, Junyang},
  booktitle={Advances in Neural Information Processing Systems},
  volume={38},
  year={2025}
}

@article{ba2016layer,
  title={Layer Normalization},
  author={Ba, Jimmy Lei and Kiros, Jamie Ryan and Hinton, Geoffrey E.},
  journal={arXiv:1607.06450},
  year={2016}
}

@inproceedings{di2023over,
  title={On Over-Squashing in Message Passing Neural Networks: The Impact of Width, Depth, and Topology},
  author={Di Giovanni, Francesco and Giusti, Lorenzo and Barbero, Federico and Luise, Giulia and Li{\`o}, Pietro and Bronstein, Michael M},
  booktitle={International Conference on Machine Learning},
  pages={7865--7885},
  year={2023}
}

@inproceedings{kipf2017semi,
  title={Semi-Supervised Classification with Graph Convolutional Networks},
  author={Kipf, Thomas N and Welling, Max},
  booktitle={International Conference on Learning Representations},
  year={2017}
}

@article{borgi2025polynomial,
  title={Polynomial Neural Sheaf Diffusion: A Spectral Filtering Approach on Cellular Sheaves},
  author={Borgi, Alessio and Silvestri, Fabrizio and Li{\`o}, Pietro},
  journal={arXiv:2512.00242},
  year={2025}
}

@inproceedings{lim2021large,
  title={Large-Scale Learning on Non-Homophilous Graphs: New Benchmarks and Strong Simple Methods},
  author={Lim, Derek and Hohne, Felix and Li, Xiuyu and Huang, Sijia Linda and Gupta, Vaishnavi and Bhalerao, Omkar and Lim, Ser Nam},
  booktitle={Advances in Neural Information Processing Systems},
  volume={34},
  pages={20887--20902},
  year={2021}
}

@article{bechler2026billion,
  title={Billion-Scale Graph Foundation Models},
  author={Bechler-Speicher, Maya and Gottlieb, Yoel and Isakov, Andrey and Abensur, David and Tavory, Ami and Haimovich, Daniel and Guy, Ido and Weinsberg, Udi},
  journal={arXiv:2602.04768},
  year={2026}
}

@inproceedings{zaghen2024sheaf,
  title={Sheaf Diffusion Goes Nonlinear: Enhancing {GNN}s with Adaptive Sheaf Laplacians},
  author={Zaghen, Olga and Longa, Antonio and Azzolin, Steve and Telyatnikov, Lev and Passerini, Andrea and Li{\`o}, Pietro},
  booktitle={Proceedings of the Geometry-grounded Representation Learning and Generative Modeling Workshop (GRaM)},
  pages={264--276},
  volume={251},
  organization={PMLR},
  year={2024}
}

@article{hansen2019toward,
  title={Toward a Spectral Theory of Cellular Sheaves},
  author={Hansen, Jakob and Ghrist, Robert},
  journal={Journal of Applied and Computational Topology},
  volume={3},
  number={4},
  pages={315--358},
  year={2019},
  publisher={Springer}
}

@book{curry2014sheaves,
  title={Sheaves, Cosheaves and Applications},
  author={Curry, Justin Michael},
  year={2014},
  publisher={University of Pennsylvania}
}

@book{bredon1997sheaf,
  title={Sheaf Theory},
  author={Bredon, Glen E},
  volume={170},
  year={1997},
  publisher={Springer Science \& Business Media}
}

@article{bourgerie2026euclidean,
  title={From Euclidean to Graph-Structured Data: A Survey of Collaborative Learning},
  author={Bourgerie, R{\'e}mi and Girdzijauskas, Sarunas and Fodor, Vikt{\'o}ria},
  journal={Transactions on Machine Learning Research},
  year={2026}
}

@article{vaswani2017attention,
  title={Attention Is All You Need},
  author={Vaswani, Ashish and Shazeer, Noam and Parmar, Niki and Uszkoreit, Jakob and Jones, Llion and Gomez, Aidan N and Kaiser, {\L}ukasz and Polosukhin, Illia},
  journal={Advances in Neural Information Processing Systems},
  volume={30},
  year={2017}
}

@inproceedings{xiao2023efficient,
  title={Efficient Streaming Language Models with Attention Sinks},
  author={Xiao, Guangxuan and Tian, Yuandong and Chen, Beidi and Han, Song and Lewis, Mike},
  booktitle={International Conference on Learning Representations},
  year={2024}
}

@inproceedings{darcet2023vision,
  title={Vision Transformers Need Registers},
  author={Darcet, Timoth{\'e}e and Oquab, Maxime and Mairal, Julien and Bojanowski, Piotr},
  booktitle={International Conference on Learning Representations},
  year={2024}
}

@inproceedings{chen2020measuring,
  title={Measuring and Relieving the Over-Smoothing Problem for Graph Neural Networks from the Topological View},
  author={Chen, Deli and Lin, Yankai and Li, Wei and Li, Peng and Zhou, Jie and Sun, Xu},
  booktitle={Proceedings of the AAAI Conference on Artificial Intelligence},
  volume={34},
  number={04},
  pages={3438--3445},
  year={2020}
}

@article{wang2025graph,
  title={Graph Foundation Models: A Comprehensive Survey},
  author={Wang, Zehong and Liu, Zheyuan and Ma, Tianyi and Li, Jiazheng and Zhang, Zheyuan and Fu, Xingbo and Li, Yiyang and Yuan, Zhengqing and Song, Wei and Ma, Yijun and others},
  journal={arXiv:2505.15116},
  year={2025}
}

@inproceedings{bechler2025position,
  title={Position: Graph Learning Will Lose Relevance Due to Poor Benchmarks},
  author={Bechler-Speicher, Maya and Finkelshtein, Ben and Frasca, Fabrizio and M{"u}ller, Luis and T{"o}nshoff, Jan and Siraudin, Antoine and Zaverkin, Viktor and Bronstein, Michael M and Niepert, Mathias and Perozzi, Bryan and Galkin, Mikhail and Morris, Christopher},
  booktitle={Proceedings of the 42nd International Conference on Machine Learning},
  pages={81067--81089},
  volume={267},
  organization={PMLR},
  year={2025}
}

@inproceedings{hanks2025distributed,
  title={Distributed Multi-Agent Coordination over Cellular Sheaves},
  author={Hanks, Tyler and Riess, Hans and Cohen, Samuel and Gross, Trevor and Hale, Matthew and Fairbanks, James},
  booktitle={IEEE Conference on Decision and Control},
  pages={3057--3064},
  year={2025}
}

@inproceedings{he2016deep,
  title={Deep Residual Learning for Image Recognition},
  author={He, Kaiming and Zhang, Xiangyu and Ren, Shaoqing and Sun, Jian},
  booktitle={Proceedings of the IEEE/CVF Conference on Computer Vision and Pattern Recognition},
  pages={770--778},
  year={2016}
}

@article{wu2023demystifying,
  title={Demystifying Oversmoothing in Attention-Based Graph Neural Networks},
  author={Wu, Xinyi and Ajorlou, Amir and Wu, Zihui and Jadbabaie, Ali},
  journal={Advances in Neural Information Processing Systems},
  volume={36},
  pages={35084--35106},
  year={2023}
}

@inproceedings{barbero2022sheafattention,
  title={Sheaf Attention Networks},
  author={Barbero, Federico and Bodnar, Cristian and de Oc{\'a}riz Borde, Haitz S{\'a}ez and Li{\`o}, Pietro},
  booktitle={NeurIPS 2022 Workshop on Symmetry and Geometry in Neural Representations},
  year={2022}
}

@inproceedings{bamberger2024bundle,
  title={Bundle Neural Networks for Message Diffusion on Graphs},
  author={Bamberger, Jacob and Barbero, Federico and Dong, Xiaowen and Bronstein, Michael M},
  booktitle={International Conference on Learning Representations},
  pages={76947--76972},
  year={2025}
}

@inproceedings{amadocomplex,
  title={Complex-Weighted Convolutional Networks: Provable Expressiveness via Complex Diffusion},
  author={Amado, Cristina L{\'o}pez and Schwarz, Tassilo and Tian, Yu and Lambiotte, Renaud},
  booktitle={The Fourth Learning on Graphs Conference},
  year={2025}
}
\bibliographystyle{icml2026}

\newpage
\appendix
\onecolumn

\section{DNSD Architecture Details}
\label{app:architecture}

We give the complete layer-by-layer specification of \textsc{DNSD}, using the notation
established in Section~\ref{sec:background}.
Recall that each node $v$ has a stalk representation
$\mathbf{X}_v^{(l)} \in \mathbb{R}^{d \times f}$,
where $d$ is the stalk dimension (number of stalks) and $f$ is the feature width per stalk,
giving total per-node representation capacity $c = d \cdot f$.

\paragraph{Input projection.}
Raw node features $\mathbf{h}_v \in \mathbb{R}^{F}$ are projected into the stalk space:
\begin{equation}
    \operatorname{vec}\!\left(\mathbf{X}_v^{(0)}\right)
    = \mathbf{W}_{\mathrm{in}}\, \mathbf{h}_v + \mathbf{b}_{\mathrm{in}},
    \qquad \mathbf{W}_{\mathrm{in}} \in \mathbb{R}^{c \times F}.
\end{equation}

\paragraph{Restriction maps.}
At each layer $l$, two independent restriction map builders compute per-edge matrices
\begin{equation}
    \mathcal{F}^{(l)}_{\mathrm{src}, e},\;
    \mathcal{F}^{(l)}_{\mathrm{tgt}, e} \in \mathbb{R}^{d \times d},
    \quad e = (u \to v),
\end{equation}
from the concatenated node embeddings:
\begin{equation}
    \mathcal{F}^{(l)}_{\bullet, e}
    = \operatorname{reshape}_{d \times d}\!\left( \phi\left(
        \mathbf{V}_{\bullet}^{(l)}\,
        \begin{bmatrix}
          \operatorname{vec}\!\left(\mathbf{X}_u^{(l)}\right) \\
          \operatorname{vec}\!\left(\mathbf{X}_v^{(l)}\right)
        \end{bmatrix}
        + \mathbf{b}^{(l)}_{\bullet}
      \right)\!\right),
    \qquad \mathbf{V}_{\bullet}^{(l)} \in \mathbb{R}^{d^2 \times 2c},
\end{equation}
where $\bullet \in \{\mathrm{src}, \mathrm{tgt}\}$ with independent parameters per layer.
Three choices of $\phi$ are considered:
\begin{itemize}
    \item \textbf{Full:} $\phi(\mathbf{z}) = \tanh(\mathbf{z})$ with $\mathbf{z} \in \mathbb{R}^{d^2}$,
          reshaped to $\mathbb{R}^{d\times d}$ by the outer reshape,
          with weights initialised at scale $1/\sqrt{2}$.
    \item \textbf{Diagonal:} $\phi(\mathbf{z}) = \mathrm{diag}(\tanh(\mathbf{z}))$ with
          $\mathbf{V}_\bullet^{(l)} \in \mathbb{R}^{d \times 2c}$ and $\mathbf{z} \in \mathbb{R}^{d}$;
          $\mathrm{diag}(\cdot)$ directly produces $\mathbb{R}^{d \times d}$, so the outer reshape is omitted.
    \item \textbf{Orthogonal:} $\phi(\mathbf{z}) = Q$ with $\mathbf{z} \in \mathbb{R}^{d^2}$,
          where the outer reshape and QR decomposition are absorbed into $\phi$:
          $\mathbf{z}$ is reshaped to $\mathbb{R}^{d\times d}$, factored as $\mathbf{z} = QR$,
          and $Q \in \mathbb{R}^{d\times d}$ is the orthogonal factor ($Q^\top Q = I$).
          The raw matrix is the learnable parameter, with $Q$ recomputed at each forward pass,
          ensuring $\mathcal{F}_{\bullet,e}^{(l)} \in O(d)$ without explicit constraints on the weights.
\end{itemize}

\paragraph{Sheaf diffusion.}
The per-edge coboundary for stalk $s \in \{1,\ldots,d\}$ is:
\begin{equation}
    \boldsymbol{\delta}_{e}^{(s,l)}
    = \bigl[\mathcal{F}^{(l)}_{\mathrm{src},e}\bigr]_{s,:}\,\mathbf{X}_u^{(l)}
      - \bigl[\mathcal{F}^{(l)}_{\mathrm{tgt},e}\bigr]_{s,:}\,\mathbf{X}_v^{(l)},
    \quad \boldsymbol{\delta}_{e}^{(s,l)} \in \mathbb{R}^{f}.
    \label{eq:coboundary_app}
\end{equation}
With the \textbf{adjacency} flag, only the target term is retained:
\begin{equation}
    \boldsymbol{\delta}_{e}^{(s,l)}
    = \bigl[\mathcal{F}^{(l)}_{\mathrm{tgt},e}\bigr]_{s,:}\,\mathbf{X}_v^{(l)}.
    \label{eq:directional_app}
\end{equation}
Messages are back-projected and symmetrically normalised:
\begin{equation}
    \left[\mathbf{m}_{u \leftarrow e}^{(l)}\right]_{s,:}
    = \tilde{d}_u^{-1/2}\,\tilde{d}_v^{-1/2}
      \sum_{j=1}^{d}
      \bigl[\mathcal{F}^{(l)}_{\mathrm{src},e}\bigr]_{js}
      \,\boldsymbol{\delta}_{e}^{(j,l)},
\end{equation}
where $\tilde{d}_v = \max(1, \deg(v))$,
and aggregated by summation:
$\bar{\mathbf{X}}_v^{(l)} = \sum_{e \ni v} \mathbf{m}_{v \leftarrow e}^{(l)} \in \mathbb{R}^{d \times f}$.
This corresponds to $\Delta_{\mathcal{F}}^{(l)}\mathbf{X}^{(l)}$ in the compact notation of~\eqref{eq:sharp}
(or $A_{\mathcal{F}}^{(l)}\mathbf{X}^{(l)}$ with the adjacency formulation).

\paragraph{Stalk-wise update.}
The aggregated signal is transformed via $W_1^{(l)} \in \mathbb{R}^{d \times d}$ and
$W_2^{(l)} \in \mathbb{R}^{f \times f}$:
\begin{equation}
    \tilde{\mathbf{X}}_v^{(l)}
    = \sigma\!\left(\bar{\mathbf{X}}_v^{(l)} \left(I_{n} \otimes W_1^{(l)}\right)
\mathbf{X}^{(l)}
W_2^{(l)}\right),
    \qquad \tilde{\mathbf{X}}_v^{(l)} \in \mathbb{R}^{d \times f},
    \label{eq:update_app}
\end{equation}
where $\sigma = \mathrm{ReLU}$ by default, or $\sigma = \tanh$ with the \textbf{odd} flag.
Both matrices are initialised as $\mathbf{I} + 0.01\,\mathcal{N}(0,1)$.

\paragraph{Stalk gating.}
With the \textbf{gate} flag, a scalar gate per stalk modulates the update.
The gate is computed from the current embedding and the aggregated signal:
\begin{equation}
    (\mathbf{G}^{(l)})_{v,s}
    = \mathrm{sigmoid}\!\left(
        w_g^{(l)}\,
        \begin{bmatrix}
          \mathbf{X}_{v,s}^{(l)} \\
          \bar{\mathbf{X}}_{v,s}^{(l)}
        \end{bmatrix}
        + b_g^{(l)}
      \right) \in [0,1],
    \qquad w_g^{(l)} \in \mathbb{R}^{1 \times 2f},
    \label{eq:gate_app}
\end{equation}
with $w_g^{(l)}$ and $b_g^{(l)}$ zero-initialised (gate $\approx 0.5$ at start).

The gated update is $(\mathbf{g}_v^{(l)} \otimes \mathbf{1}_f^\top) \odot \tilde{\mathbf{X}}_v^{(l)}$,
consistent with the compact form in~\eqref{eq:sharp}.

\paragraph{Residual and layer normalisation.}
Following~\eqref{eq:sharp}, the full layer update is:
\begin{equation}
    \mathbf{X}_v^{(l+1)}
    = \mathrm{LN}\!\left(
        (1 + \epsilon^{(l)})\,\mathbf{X}_v^{(l)}
        -
        (\mathbf{G}^{(l)})_v \otimes \mathbf{1}_f^\top) \odot \tilde{\mathbf{X}}_v^{(l)}
      \right),
\end{equation}
where $\epsilon^{(l)} \in \mathbb{R}$ is a learnable scalar (zero-initialised) and
$\mathrm{LN}$ is layer normalisation applied per stalk as in~\eqref{eq:layernorm}, normalising across the feature dimension $f$ independently for each stalk.

\paragraph{Output classifier.}
\begin{equation}
    \hat{\mathbf{y}}_v
    = \mathbf{W}_{\mathrm{out}}\,\operatorname{vec}\!\left(\mathbf{X}_v^{(L)}\right)
      + \mathbf{b}_{\mathrm{out}},
    \qquad \mathbf{W}_{\mathrm{out}} \in \mathbb{R}^{C \times c},
\end{equation}
where $C$ represents the number of classes.
\paragraph{Summary of flags.} Table \ref{tab:flags} summarizes the \textsc{DNSD} architecture options and their realizations on the base NSD equation.

\begin{table}[h!]
\centering
\caption{\textsc{DNSD} architectural flags and their effect on the base update.}
\label{tab:flags}
\small
\begin{tabular}{lll}
\toprule
Flag & Off (default) & On \\
\midrule
\texttt{adj}   & Symmetric coboundary (Eq.~\ref{eq:coboundary_app})
               & Target-only (Eq.~\ref{eq:directional_app}) \\
\texttt{odd}   & $\sigma = \mathrm{ReLU}$ in Eq.~\ref{eq:update_app}
               & $\sigma = \tanh$ \\
\texttt{gate} & $\mathbf{g}_v^{(l)} \equiv \mathbf{1}_d$
               & Per-stalk sigmoid gate (Eq.~\ref{eq:gate_app}) \\
\bottomrule
\end{tabular}
\end{table}

\section{Experimental Details}
\label{app:details}

In this section, we give further details about the experiment settings on the synthetic and real-world benchmark.

\subsection{Synthetic Community Detection Benchmark}
\label{app:experiments_synthetic}

\paragraph{Benchmark Construction.}
We construct a synthetic community detection benchmark based on a $K$-nearest-neighbour (KNN) graph with $N=1{,}500$ nodes partitioned into $K=3$ communities of equal size (500 nodes each). Node features are 2-dimensional, sampled from Gaussian distributions centred at community-specific means with standard deviation $\sigma=3.0$. The KNN graph is built with $k=8$ neighbours; the intra-community edge probability of an equivalent SBM is set to match the average KNN degree. The heterophily level $L\in\{0,1,\ldots,10\}$ controls the degree of inter-community mixing via edge perturbation: at $L=0$ the graph is fully homophilic, while at $L=10$ edges are entirely inter-community. Train and test graphs are generated independently from disjoint random seeds to avoid data leakage and cached to disk for reproducibility. The train graph uses an 80/20 node-level train/validation split; evaluation is performed on the independently generated test graph over all nodes.

\paragraph{Dataset statistics.}
Each benchmark graph has $N=1{,}500$ nodes, ${\approx}7{,}257$ edges, $F=2$ node features, and $C=3$ classes. The edge count is constant across levels since perturbation rewires edges rather than adding or removing them.

\paragraph{Hyperparameters.}
We use the Adam optimiser with learning rate $\eta=0.01$ and weight decay $\lambda=5\times10^{-4}$. The learning rate is reduced on plateau (factor 0.5, patience 20 epochs). The hidden dimension is fixed at $c=18$ and the number of stalks at $d=3$ (so $f=6$, equal to $K$). We perform a grid search over the number of layers $\in\{2,4,8,12,16\}$. Training runs for a maximum of 500 epochs with early stopping (patience 100) monitored on validation accuracy; the best checkpoint is restored at the end of training. Results are reported as mean\,$\pm$\,std over 6 random train seeds $\{42,43,44,45,46,47\}$, evaluated on test graphs generated from 3 independent test seeds $\{100,101,102\}$.

\paragraph{Model complexity.}
Table~\ref{tab:params_syn} reports parameter counts at each model's selected depth
(chosen by mean validation accuracy across all perturbation levels and training seeds).
All models use $c{=}18$, $d{=}3$, $f{=}6$, and $F{=}2$ (2D Gaussian node features).

\begin{table}[h]
\centering
\caption{Parameter counts for synthetic benchmark models ($c{=}18$, $d{=}3$, $f{=}6$, node features $F{=}2$, classes $C{=}3$). Total counts shown for a representative graph; backbone excludes input/output projections. }
\label{tab:params_syn}
\small
\begin{tabular}{llcrr}
\toprule
Model & Flags & $L$ & Total & Backbone \\
\midrule
MLP          & -- & 2  &    111 & -- \\
GAT          & -- & 4  &    909 & -- \\
\midrule
NSD diag     & -- & 16 &  2{,}655 & 2{,}544 \\
NSD full     & -- &  8 &  3{,}159 & 3{,}048 \\
\midrule
\textsc{DNSD} diag  & --                          & 16 &  5{,}007 & 4{,}896 \\
             & \texttt{adj}                & 16 &  5{,}007 & 4{,}896 \\
             & \texttt{odd}                & 16 &  5{,}007 & 4{,}896 \\
             & \texttt{gate}              & 16 &  5{,}215 & 5{,}104 \\
             & \texttt{adj+odd}            & 16 &  5{,}007 & 4{,}896 \\
             & \texttt{adj+gate}          & 12 &  3{,}939 & 3{,}828 \\
             & \texttt{odd+gate}          & 16 &  5{,}215 & 5{,}104 \\
             & \texttt{adj+odd+gate}      & 12 &  3{,}939 & 3{,}828 \\
\midrule
\textsc{DNSD} full  & --                          & 16 & 12{,}111 & 12{,}000 \\
             & \texttt{adj}                & 16 & 12{,}111 & 12{,}000 \\
             & \texttt{odd}                & 16 & 12{,}111 & 12{,}000 \\
             & \texttt{gate}              & 16 & 12{,}319 & 12{,}208 \\
             & \texttt{adj+odd}            & 16 & 12{,}111 & 12{,}000 \\
             & \texttt{adj+gate}          & 16 & 12{,}319 & 12{,}208 \\
             & \texttt{odd+gate}          & 16 & 12{,}319 & 12{,}208 \\
             & \texttt{adj+odd+gate}      & 16 & 12{,}319 & 12{,}208 \\
\bottomrule
\end{tabular}
\end{table}

\paragraph{Computational resources.}
Experiments were run on a standard CPU workstation over approximately two weeks of wall-clock time.

\subsection{Real-World Benchmark}
\label{app:real_world_experiments}

\paragraph{Datasets.}
We evaluate on 9 real-world node classification benchmarks: Actor, Chameleon (filtered), and Squirrel (filtered), roman-empire, amazon-ratings, minesweeper, tolokers, and questions~\citep{platonov2023critical}, and penn94~\citep{lim2021large}. The filtered Chameleon and Squirrel variants remove duplicate nodes following~\citep{platonov2023critical}.

\paragraph{Dataset statistics.}
Table~\ref{tab:dataset_stats} summarises the graph statistics. Heterophily is measured as the fraction of edges connecting nodes of different classes.

\begin{table}[h]
\centering
\caption{Statistics of the real-world benchmark datasets.}
\label{tab:dataset_stats}
\begin{tabular}{@{}llrrrrl@{}}
\toprule
\textbf{Dataset} & \textbf{Size} & \textbf{\#Nodes} & \textbf{\#Edges} & \textbf{\#Features} & \textbf{\#Classes} & \textbf{Heterophily} \\
\midrule
Chameleon (filtered)   & small  &    890    &   4,427    & 2,325 & 5  & 0.764 \\
Squirrel (filtered)    & small  &  2,223    &  23,499    & 2,089 & 5  & 0.793 \\
\midrule
Actor                  & medium &  7,600    &  15,009    &   932 & 5  & 0.781 \\
Minesweeper            & medium & 10,000    &  39,402    &     7 & 2  & 0.317 \\
Tolokers               & medium & 11,758    & 519,000    &    10 & 2  & 0.405 \\
Roman Empire           & medium & 22,662    &  32,927    &   300 & 18 & 0.953 \\
Amazon Ratings         & medium & 24,492    &  93,050    &   300 & 5  & 0.620 \\
\midrule
Questions              & large  & 48,921    & 153,540    &   301 & 2  & 0.160 \\
Penn94                 & large  & 41,554    & 1,362,229  & 4,814 & 2  & 0.530 \\
\bottomrule
\end{tabular}
\end{table}

\paragraph{Data splits.}
All datasets use a fixed stratified random split of 60/20/20 (train/validation/test), stratified per class. Splits are generated with a dataset-name-derived fixed seed and cached to disk for reproducibility across runs.

\paragraph{Models.}
We evaluate \textsc{DNSD} with diagonal restriction maps across 8 variants obtained by combining three binary flags: adjacency operator (\texttt{adj}), odd activation (\texttt{odd}, tanh instead of ReLU), and per-stalk gating (\texttt{gate}). We also include NSD v1 (full and diagonal), MLP, and GAT as baselines.

\paragraph{Hyperparameters.}
We perform a grid search over the number of layers $\in \{2, 4, 8\}$, with fixed hidden dimension $c=64$, number of stalks $d=8$ (so $f=8$), learning rate $\eta=0.01$, and weight decay $\lambda=5\times10^{-4}$. The maximum number of training epochs is 500. Best hyperparameters are selected per dataset--model combination based on validation accuracy.

\paragraph{Training.}
We use the Adam optimiser. The learning rate is reduced on plateau (factor 0.5, patience 20 epochs). Early stopping is applied with a patience of 100 epochs monitored on validation accuracy; the best checkpoint is restored at the end of training. All results are reported as mean $\pm$ std over 5 random seeds $\{42, 43, 44, 45, 46\}$. MPNN was excluded from the Penn94 benchmark due to memory overflow on the full graph.

\paragraph{Results.}
Table~\ref{tab:realworld} reports test accuracy (\%) per model and dataset. Each entry shows the mean\,$\pm$\,std across seeds at the best layer depth for that (dataset, model) combination, selected by mean validation accuracy; the optimal depth is shown in parentheses. \textsc{DNSD} variants with diagonal restriction maps consistently outperform NSD v1 and standard GNN baselines on heterophilic datasets, with the adjacency flag (\texttt{adj}) providing the most consistent gains across datasets. On homophily-insensitive datasets such as Questions, the differences between the models are minimal.

\begin{table}[tpb]
\centering
\caption{Node classification accuracy (\%) on heterophilic benchmarks. All \textsc{DNSD} variants use diagonal restriction maps and include LayerNorm. Each entry reports mean\,$\pm$\,std at the best layer depth (in parentheses) for that (dataset, model) pair. \rone{1st}, \rtwo{2nd}, \rthree{3rd} best per column.}
\label{tab:realworld_appendix}
\resizebox{\textwidth}{!}{%
\begin{tabular}{@{}l l ccc ccccccccc@{}}
\toprule
& & \multicolumn{3}{c}{\textbf{Fixes}} & \multicolumn{9}{c}{\textbf{Dataset}} \\
\cmidrule(lr){3-5} \cmidrule(lr){6-14}
& \textbf{Map}
& \rotatebox{60}{Adj} & \rotatebox{60}{Odd} & \rotatebox{60}{Gate}
& \rotatebox{60}{Actor}
& \rotatebox{60}{\makecell{Chameleon\\(filtered)}}
& \rotatebox{60}{\makecell{Squirrel\\(filtered)}}
& \rotatebox{60}{\makecell{Roman\\Empire}}
& \rotatebox{60}{\makecell{Amazon\\Ratings}}
& \rotatebox{60}{Minesweeper}
& \rotatebox{60}{Tolokers}
& \rotatebox{60}{Questions}
& \rotatebox{60}{Penn94} \\
\midrule
MLP  & --- & & & &
\rone{37.0\sd{0.3}{\scriptsize(G4)}} & \rthree{43.0\sd{3.7}{\scriptsize(G4)}} & \rtwo{38.3\sd{0.6}{\scriptsize(G4)}} & 66.4\sd{0.1}{\scriptsize(G2)} & 40.9\sd{0.4}{\scriptsize(G2)} & 80.0\sd{0.0}{\scriptsize(G2)} & 78.2\sd{0.0}{\scriptsize(G2)} & 97.0\sd{0.0}{\scriptsize(G2)} & 76.2\sd{0.0}{\scriptsize(G2)} \\
MPNN & --- & & & &
34.4\sd{1.1}{\scriptsize(G2)} & \rone{45.5\sd{2.2}{\scriptsize(G4)}} & \rone{43.1\sd{0.5}{\scriptsize(G2)}} & 78.9\sd{0.8}{\scriptsize(G2)} & 46.6\sd{0.3}{\scriptsize(G4)} & 87.4\sd{1.3}{\scriptsize(G4)} & 79.1\sd{0.1}{\scriptsize(G2)} & 97.0\sd{0.0}{\scriptsize(G2)} & --- \\
GAT  & --- & & & &
30.3\sd{0.6}{\scriptsize(G2)} & \rtwo{45.2\sd{2.7}{\scriptsize(G4)}} & \rthree{37.0\sd{1.7}{\scriptsize(G8)}} & 56.9\sd{1.1}{\scriptsize(G2)} & 46.0\sd{0.6}{\scriptsize(G2)} & 80.3\sd{0.0}{\scriptsize(G2)} & 78.4\sd{0.1}{\scriptsize(G4)} & 97.0\sd{0.0}{\scriptsize(G2)} & 74.1\sd{2.0}{\scriptsize(G2)} \\
\midrule
\multirow{2}{*}{NSD}
& diag & & & &
\rthree{36.2\sd{0.6}{\scriptsize(G2)}} & 41.5\sd{1.2}{\scriptsize(G2)} & 35.4\sd{1.5}{\scriptsize(G2)} & 79.1\sd{0.5}{\scriptsize(G8)} & 44.6\sd{0.3}{\scriptsize(G8)} & 87.5\sd{0.6}{\scriptsize(G8)} & 81.5\sd{0.2}{\scriptsize(G8)} & 97.1\sd{0.0}{\scriptsize(G2)} & 76.3\sd{0.1}{\scriptsize(G2)} \\
& full & & & &
35.9\sd{0.3}{\scriptsize(G2)} & 41.6\sd{2.7}{\scriptsize(G4)} & 36.5\sd{1.6}{\scriptsize(G4)} & 77.1\sd{0.9}{\scriptsize(G4)} & 45.4\sd{0.7}{\scriptsize(G4)} & 86.1\sd{0.3}{\scriptsize(G4)} & 81.4\sd{0.3}{\scriptsize(G2)} & 97.1\sd{0.0}{\scriptsize(G2)} & 76.1\sd{0.6}{\scriptsize(G4)} \\
\midrule
\multirow{5}{*}{\rotatebox{90}{\textsc{DNSD} (diag)}}
& diag & \cmark & \cmark & \cmark &
\rthree{36.2\sd{0.6}{\scriptsize(G4)}} & 37.7\sd{1.1}{\scriptsize(G8)} & 33.5\sd{1.3}{\scriptsize(G8)} & \rtwo{83.4\sd{0.9}{\scriptsize(G8)}} & \rtwo{47.8\sd{0.4}{\scriptsize(G8)}} & \rone{89.4\sd{0.8}{\scriptsize(G8)}} & \rtwo{81.8\sd{0.5}{\scriptsize(G4)}} & 97.1\sd{0.0}{\scriptsize(G4)} & \rtwo{78.7\sd{0.9}{\scriptsize(G8)}} \\
& diag & \cmark & \cmark &  &
36.0\sd{0.7}{\scriptsize(G2)} & 37.9\sd{0.9}{\scriptsize(G8)} & 32.8\sd{1.0}{\scriptsize(G4)} & \rthree{83.2\sd{0.7}{\scriptsize(G8)}} & \rone{49.1\sd{0.7}{\scriptsize(G8)}} & \rtwo{88.9\sd{0.4}{\scriptsize(G8)}} & \rone{82.0\sd{0.2}{\scriptsize(G8)}} & 97.1\sd{0.0}{\scriptsize(G2)} & \rone{80.0\sd{0.9}{\scriptsize(G8)}} \\
& diag & \cmark &  & \cmark &
\rthree{36.2\sd{0.6}{\scriptsize(G4)}} & 37.9\sd{0.9}{\scriptsize(G4)} & 33.1\sd{1.2}{\scriptsize(G8)} & 83.0\sd{0.4}{\scriptsize(G8)} & \rthree{47.5\sd{1.0}{\scriptsize(G4)}} & 88.1\sd{0.8}{\scriptsize(G8)} & \rthree{81.8\sd{0.7}{\scriptsize(G8)}} & 97.1\sd{0.0}{\scriptsize(G4)} & \rthree{78.6\sd{0.9}{\scriptsize(G8)}} \\
& diag &  & \cmark & \cmark &
\rtwo{36.4\sd{0.4}{\scriptsize(G4)}} & 38.0\sd{0.6}{\scriptsize(G4)} & 33.5\sd{1.3}{\scriptsize(G8)} & \rone{83.4\sd{0.2}{\scriptsize(G8)}} & 46.0\sd{0.4}{\scriptsize(G2)} & \rthree{88.2\sd{0.5}{\scriptsize(G8)}} & 81.5\sd{0.4}{\scriptsize(G8)} & \rone{97.2\sd{0.0}{\scriptsize(G2)}} & 76.0\sd{0.6}{\scriptsize(G4)} \\
& diag &  &  &  &
36.1\sd{0.4}{\scriptsize(G2)} & 39.1\sd{0.9}{\scriptsize(G8)} & 33.1\sd{1.4}{\scriptsize(G8)} & 82.7\sd{0.5}{\scriptsize(G8)} & 46.8\sd{0.5}{\scriptsize(G8)} & 87.6\sd{0.6}{\scriptsize(G8)} & 81.1\sd{0.4}{\scriptsize(G2)} & \rtwo{97.2\sd{0.0}{\scriptsize(G8)}} & 76.5\sd{0.4}{\scriptsize(G8)} \\
\bottomrule
\end{tabular}%
}
\end{table}

\paragraph{Computational resources.}
Experiments were run using a node on the Dardel HPC cluster at PDC Center for High Performance Computing\footnote{https://www.pdc.kth.se/~}, equipped with 8 AMD MI250X GPUs, for a total of 70 GPU-core hours.

\section{Sheaf Diffusion and Attention}
\label{app:attention}

In this section, we make explicit the connection between sheaf-based message passing and attention mechanisms by formalizing the edge-dependent functions and update operators used in NSD and \textsc{DNSD}. We use the term graph attention mechanisms (GATs) to refer to the family of models analyzed in~\citet{brody2021attentive}, which includes Graph Attention (GAT)~\citep{velivckovic2017graph}, its more expressive variants (GATv2)~\citep{brody2021attentive}, and dot-product attention~\citep{vaswani2017attention}.

\paragraph{Edge-dependent functions.}
Both graph attention mechanisms and \textsc{DNSD} compute edge-dependent functions from pairs of node representations. At layer $l$, for a node $u$ and its neighbor $v$, this takes the form
\[
[\mathbf{A}_{\mathcal{F}}^{(l)}]_{uv} = \Phi^{(l)}\!\left(\mathbf{X}_u^{(l)}, \mathbf{X}_v^{(l)}\right),
\]
where $\Phi^{(l)}$ is a learnable function of the pair $(\mathbf{X}_u^{(l)}, \mathbf{X}_v^{(l)})$.

In graph attention mechanisms, $\Phi^{(l)}$ outputs a scalar attention score
\[
e_{uv}^{(l)} = \Phi^{(l)}\!\left(\mathbf{X}_u^{(l)}, \mathbf{X}_v^{(l)}\right) \in \mathbb{R},
\]
which is then normalized across neighbors using a softmax:
\[
\alpha_{uv}^{(l)} = \frac{\exp(e_{uv}^{(l)})}{\sum_{w \in \mathcal{N}(u)} \exp(e_{uw}^{(l)})}.
\]
The normalized coefficients $\alpha_{uv}^{(l)}$ are used to weight neighboring representations.

In contrast, \textsc{DNSD} outputs a linear map:
\[
\mathbf{A}_{\mathcal{F},uv}^{(l)} \in \mathbb{R}^{d \times d},
\]
which is applied directly to the representation $\mathbf{X}_v^{(l)}$.
\paragraph{Remark.}
The stalk dimension in \textsc{DNSD} can be viewed as playing a role loosely analogous to multiple attention heads, in that both enable multiple parallel interactions. However, the underlying mechanisms differ, as \textsc{DNSD} applies linear maps rather than scalar weightings.
\paragraph{Update operators.}
The distinction between NSD and \textsc{DNSD} lies in the operator used to compute updates. NSD applies the sheaf Laplacian:
\[
\Delta_{\mathcal{F}} \mathbf{X},
\]
which aggregates \emph{differences} between neighboring node representations. In contrast, \textsc{DNSD} uses the sheaf adjacency operator:
\[
\mathbf{A}_{\mathcal{F}} \mathbf{X},
\]
which aggregates \emph{dependencies} between neighbors.

Thus, NSD and \textsc{DNSD} differ in the signal used for aggregation, while \textsc{DNSD} and attention share a dependency-based formulation.

\paragraph{Aggregation mechanisms.}
Given edge-dependent functions, the update at node $u$ takes the form
\[
\sum_{v \in \mathcal{N}(u)} [\mathbf{A}_{\mathcal{F}}^{(l)}]_{uv} \mathbf{X}_v^{(l)}.
\]
In attention-based models, scalar attention scores are normalized via softmax, producing convex combinations of neighboring representations. In \textsc{DNSD}, no normalization is applied to the aggregation operators, and updates are not restricted to convex combinations.

\paragraph{Normalization.}
Attention mechanisms normalize scalar attention scores using softmax:
\[
\alpha_{uv}^{(l)} = \frac{\exp(e_{uv}^{(l)})}{\sum_{w \in \mathcal{N}(u)} \exp(e_{uw}^{(l)})}.
\]
In \textsc{DNSD}, normalization is instead applied to node representations after aggregation via LayerNorm. While \textsc{DNSD} inherits operator normalization from the sheaf construction, the dominant normalization occurs at the level of node representations.

\paragraph{Summary.}
The comparison can be summarized along three axes: (i) the update operator (difference vs.\ dependency), (ii) the form of edge-dependent functions (scalar scores vs.\ linear maps), and (iii) the location of normalization (scores vs.\ representations). These structural differences explain the distinct behavior observed at depth in the main text.

\end{document}